\documentclass{article} 
\usepackage{iclr2024_conference,times}

\usepackage{amsmath,amsfonts,bm}

\def\eqref#1{equation~\ref{#1}}

\def\1{\bm{1}}

\DeclareMathAlphabet{\mathsfit}{\encodingdefault}{\sfdefault}{m}{sl}
\SetMathAlphabet{\mathsfit}{bold}{\encodingdefault}{\sfdefault}{bx}{n}

\usepackage{hyperref}
\usepackage{url}
\usepackage{graphicx}
\usepackage{tcolorbox}
\tcbuselibrary{skins}

\usepackage{listings}
\usepackage{authblk}
\usepackage{xcolor}

\lstdefinelanguage{JSON}{
  basicstyle=\normalfont\ttfamily,
  numbers=left,
  numberstyle=\small,
  stepnumber=1,
  showstringspaces=false,
  breaklines=false,
  frame=lines
}

\definecolor{lightgray}{gray}{0.95}
\lstdefinestyle{myprompt}{
    basicstyle=\ttfamily\fontsize{7pt}{8pt}\selectfont,
    frame=none,
    breaklines=true,
    backgroundcolor=\color{lightgray},
    breakatwhitespace=true,
    breakindent=0pt,
    escapeinside={(*@}{@*)},
    numbers=none,
    numbersep=5pt,
    xleftmargin=5pt,
}
\tcbset{
  myaibox/.style={
    top=10pt,
    colback=white,
    colframe=black,
    colbacktitle=black,
    enhanced,
    center,
    attach boxed title to top left={yshift=-0.1in,xshift=0.15in},
    boxed title style={boxrule=0pt,colframe=white,},
  }
}
\newtcolorbox{AIbox}[2][]{myaibox,title=#2,#1}

\title{labeling supervised fine-tuning data with the scaling law}

\author[1,2]{Huanjun Kong}

\affil[1]{ncnn contributors}
\affil[2]{Shanghai AI Laboratory}

\begin{document}

\maketitle

\begin{abstract}
This paper introduces a multi-stage manual annotation calibrated by the scaling law, \textbf{offering a high-quality Supervised Fine-Tuning data acquisition method for environments with constrained resources} like GPU poor, limited GPT access, and funding restrictions. We have preprocessed 58k authentic chat data and manually annotated 2.3k questions. After this, we conducted fine-tuning on Qwen models, ranging from 0.5B to 32B parameters. The optimal version improved 29.07 in F1 score. This confirms the viability of fine-tuning Large Language Model (LLM) for downstream Natural Language Processing (NLP) tasks. Our contributions are: 1) Created Supervised Fine-Tuning (SFT) training data in alpaca format, along with a set of Low-Rank Adaptation (LoRA) weights, and 2) Developed a method for acquiring high-quality data leveraging scaling law principle. The script, raw data with alpaca format and experiments track are open-sourced on \href{https://github.com/InternLM/HuixiangDou/tree/main/web/tools}{Github}, \href{https://huggingface.co/tpoisonooo}{HuggingFace} and \href{https://wandb.ai/tpoisonooo/huixiangdou-cr/table?nw=nwusertpoisonooo}{WandB}. The privacy of the data involved has been authorized by users\footnote{SFT data and license comes from ncnn contributors group.}.
\end{abstract}

\section{Introduction and Related Work}

In HuixiangDou \cite{kong2024huixiangdou}, we have released an industrial-grade assistant to handle inquiries within group chats. We also build rejection pipeline with text2vec model and LLM scoring. However, ChatML format is limited to the roles of \textbf{bot}, \textbf{system}, and \textbf{user}, with prompts in HuixiangDou only consisted of individual user inputs, which does not take full advantage of the dialogue from other users. This limitation is particularly evident when dealing with conversation involving pronouns, as shown in conversation example \ref{fig:conversation}.
\newline
\begin{figure*}[h]
\vspace{-5mm}
\begin{AIbox}{Group Chat Example}
{\bf Alice:} \\
{
Can \textcolor{blue}{mmpose} be deployed on mobile phones?
}\\ 
{\bf Bob:} \\
{
BTW, how to deploy \textcolor{blue}{it} on TX2 ?
}
\end{AIbox} 
\caption{A conversational example containing pronouns, in which "it" needs to be disambiguated to "mmpose".}
\label{fig:conversation}
\end{figure*}

We have experimented NLP methods in HuixiangDou, but apart from the straightforward segmentation of a single question into multiple parts using part-of-speech (POS) tagging, other methods have proven to be of low cost-effectiveness. On one hand, the NLP implementation is complex; on the other hand, open-sourced models like HanLP \cite{he-choi-2021-stem} are unable to support both Chinese and English well. Additionally, the part-of-speech in Chinese is more complex than in English. For instance, the Chinese equivalent of the word "deploy" can function as a verb, adjective, or noun depending on the context. When faced with thousands of users, there are always badcases that break NLP rules, leading to extremely high maintenance costs for the system.

ReALM \cite{moniz2024realm} distinguishes three coreference resolution task types in mobile scenario, claiming that with manual annotation and training, a fine-tuned model with 80 million parameters can rival the capabilities of GPT-4. \cite{qin2022exploring} attempts to address how Pre-trained Language Models (PLMs) can learn universal representations and effectively adapt to a wide range of NLP tasks. Therefore, we believe that fine-tuning LLM for downstream NLP tasks is feasible.

\cite{li2024quantity} introduced Instruction-Following Difficulty (IFD) and has demonstrated that by filtering out 5\% of the data, then better accuracy can be achieved. It proves that the quality of SFT data is more important than quantity. This raises the question for us: \textbf{what is the objective criteria for data quality in this scenario?}

Both LoRA \cite{hu2021lora} and LoRA+ \cite{hayou2024lora} assume that the increments of weights are of low rank. According to inferences made on \cite{tpoisonooo662295224}, LoRA only activates the pre-existing knowledge in PLM but can't learn new content. So, how can we prove that the data in this scenario meets the low-rank assumption?

\section{Methodology}
\label{label:methodology}

HuixiangDou has been active in many groups for a year. Given the diverse backgrounds of the ncnn contributors and the high complexity of the topics discussed, which range from hardware to software and possess a certain degree of universality, and the group's high level activity—87 messages per person per month on average—makes it ideal for experiments.

Consequently, we have preprocessed 58,000 raw chat lines from ncnn contributors group, following these two steps:

\begin{itemize}
    \item \textbf{Concat} Concatenating consecutive messages from the same user based on timestamp. In post-hoc analysis, a person need to send 2 messages to articulate 1 question.
    \item \textbf{Filter} Utilizing KIMI and InternLM2 to assess whether a query is a question, since HuixiangDou is primarily concerned inquiry within the group.
\end{itemize}

After preprocess is completed, we have 2,302 inquiries.

Then we take scaling law as objective metric for manual annotation. \textbf{The larger the model's parameter, the better vanilla LLM performance on the annotated data will be.}

\begin{figure}[ht]
\centering
\includegraphics[width=0.2\textwidth]{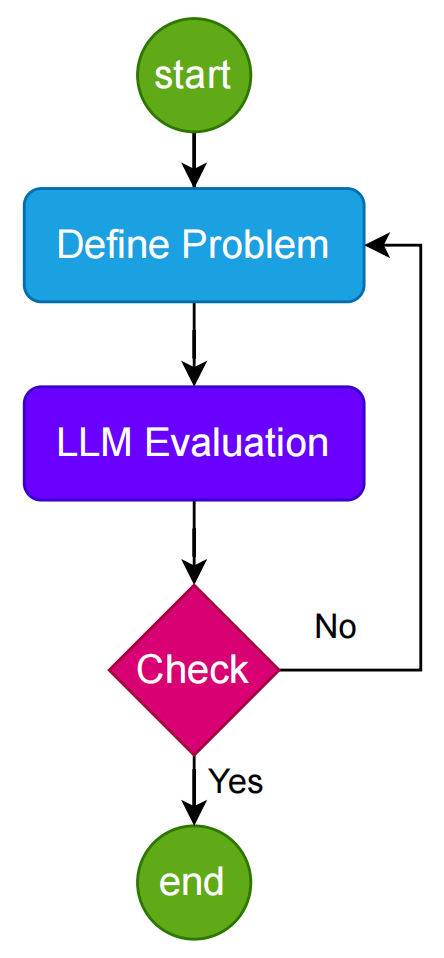}
\caption{\label{fig:annotation}The overall structure of data annotation steps.}
\end{figure}

As shown in Figure \ref{fig:annotation}, our annotation process involved the following steps:

\begin{enumerate}
    \item \textbf{Problem Definition} Describe the task,  manually annotate and define LLM prompt. 
    \item \textbf{LLM Evaluation} Take 7 Qwen LLMs (from 0.5B to 32B) as baseline and test them on the annotated data.
    \item \textbf{Check} If precision does not increase with model size expanding, then revise the prompt and re-annotate, goto step 1.
\end{enumerate}

After several iterations, the precision of the baseline models is shown in Table \ref{tab:llm_baseline}.

\begin{table}[h]
    \centering
    
\begin{tabular}{lcc}
\hline
\textbf{Model} & \textbf{Precision(\%)} & \textbf{F1 score(\%)}\\\hline
Qwen1.5-0.5B-Chat & \textbf{\color{blue}55.68} & 52.43 \\
Qwen1.5-1.8B-Chat & \textbf{\color{blue}56.15} & 50.02 \\
Qwen1.5-4B-Chat & \textbf{\color{blue}56.92} & 62.75 \\
Qwen1.5-7B-Chat & \textbf{\color{blue}57.11} & 68.17 \\
Qwen1.5-14B-Chat & \textbf{\color{blue}60.77} & 64.59 \\
Qwen1.5-32B-Chat & \textbf{\color{blue}68.29} & 60.15 \\
Qwen1.5-MoE-2.7B-Chat & 61.79 & 32.86 \\\hline

\end{tabular}
\caption{Qwen LLMs precision is tested on annotated data. Based on the objectivity of the scaling law, the blue part confirms the reliability of manual annotation. With the change in model architecture, Mixture of Experts (MoE) result is only for reference.}
\label{tab:llm_baseline}
\end{table}

We fine-tuned 7 models on annotated data. Ultimately, Qwen1.5-MoE-2.7B-Chat F1 score significantly improved from 32.07\% to 61.93\%. Moreover, Qwen1.5-14B-Chat excelled in recall increase, from 68.91\% to 92.11\%.

\section{Dataset}
\label{label:dataset}

Our preprocessed group chat data with JSON format is shown below, with paramater explaination:

\begin{lstlisting}[language=JSON]
{
  "id": INTEGER,
  "sender": "STRING",
  "text": "SRING",
  "timestamp": INTEGER,
  "kimi_is_question": BOOL,
  "cr_window": [{
    "sender": "STRING",
    "text": "STRING",
    "timestamp": INTEGER
  }],
  "cr_need_gt": BOOL
}

# sender: message sender ID, it is random STRING, not username
# is_question: filter by InternLM2 scoring, does this text is 
a question
# kimi_is_question: filter by KIMI scoring, the throttle is 5
# cr_window: historical messages before the text, up to a maximum of 12
# cr_need_gt: manual annotation, does this text need resolution or not
\end{lstlisting}

As alpaca format below needs \textbf{input} or \textbf{instruction}, we have to merge \textbf{text} and \textbf{cr\_window} into single sentence.

\begin{lstlisting}[language=JSON]
{
  "instruction": "STRING",
  "input": "STRING",
  "output": "STRING"
}
\end{lstlisting}

In this scenario, even if coreference resolution fails, the original input is retained, allowing LLM making ternary  decision to disambiguate the input, thus alpaca \textbf{instruct} includes single-choice question formats, such as "A. Not needed B. Needed C. Don't know".

\paragraph{Implementation Details} To prevent overfitting, we randomly adjusted the correct option position in alpaca \textbf{output}, with some data having the true value in option A and others in option B or C, and the corresponding implementations are available on \href{https://github.com/InternLM/HuixiangDou/blob/main/web/tools/convert_to_alpaca.py#L4}{Github}.

\section{SFT Experiments}
\label{label:training}

We fine-tuned the Qwen series models using LoRA in axolotl \cite{Axolotl2023}, all configurations can be found on \href{https://github.com/InternLM/HuixiangDou}{Github}.

At the initial state, we set epoch to 1, learning rate to 2e-4 and LoRA rank to 64. The accuracy is detailed in Table \ref{tab:llm_1}.

\begin{table}[h]
\centering
\begin{tabular}{lcc}
\hline
\textbf{Model} & \textbf{F1 score(\%)} & \textbf{improvement(\%)} \\\hline
Qwen1.5-0.5B-Chat & \color{blue}60.60 & +8.17 \\
Qwen1.5-1.8B-Chat & \color{blue}62.90 & +12.88 \\
Qwen1.5-4B-Chat & \color{blue}51.25 & \color{red}-11.50 \\
Qwen1.5-7B-Chat & \color{blue}69.22 & +1.05 \\
Qwen1.5-14B-Chat & \color{blue}77.09 & +12.50 \\
Qwen1.5-32B-Chat & \color{blue}85.58 & +25.43 \\
Qwen1.5-MoE-2.7B-Chat & 61.93 & \color{teal}+29.07 \\\hline

\end{tabular}
\caption{Finetuned Qwen models F1 score, the blue part also exhibits a trend in line with the scaling law, which validates the reliability of the annotation.}
\label{tab:llm_1}
\end{table}

\begin{figure}[ht]
\centering
\includegraphics[width=0.8\textwidth]{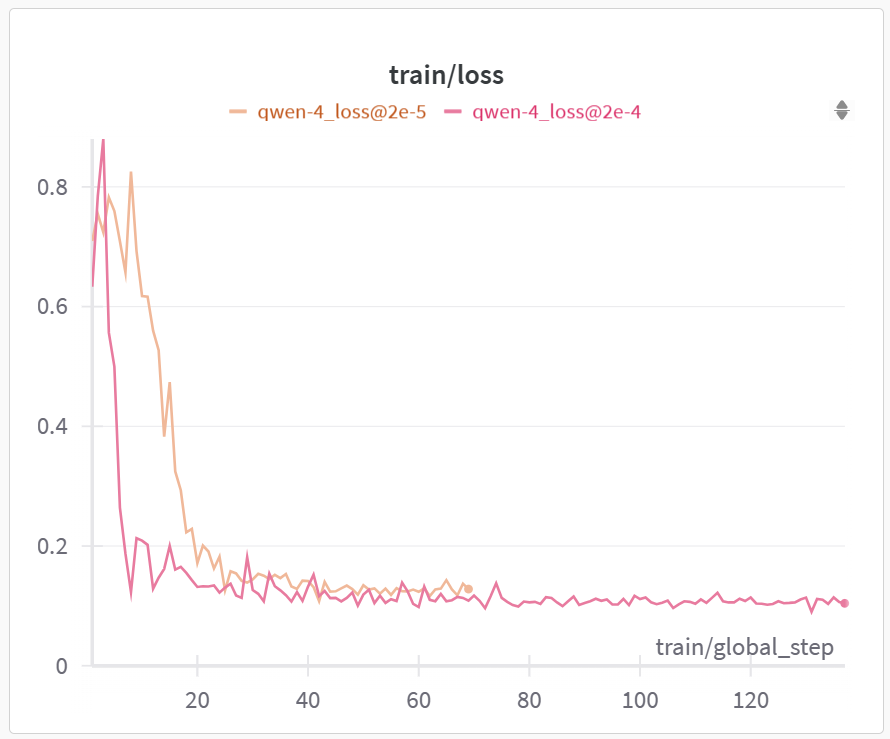}
\caption{\label{fig:qwen4-loss}
The curves representing different learning rates indicate that when Qwen1.5-4B-Chat learning rate set to 2e-5, the loss in F1 score is mitigated.}
\end{figure}

We observed a F1 score drop in Qwen1.5-4B-Chat model and we believe that a rapid decrease in loss might affect performance, so we updated the learning rate from 2e-4 to 2e-5. This adjustment effectively mitigated the decline in F1 score, which improved from -11.50 to -3.44. The new loss curve shown in Figure \ref{fig:qwen4-loss}.

We also experimented with various rank size and LoRA+ for optimization, the results are presented in Table \ref{tab:llm_4B}.

\begin{table}[h]
\centering
\begin{tabular}{lcc}
\hline
\textbf{Method} & \textbf{F1 score(\%)} \\\hline
vanilla & 62.75 \\
LoRA lr=2e-5 & 59.31 \\
LoRA lr=2e-4 & 51.25 \\
LoRA rank=16 & 53.27 \\
LoRA rank=32 & 56.54 \\
LoRA rank=64 & 51.25 \\ 
LoRA rank=128 & 49.31 \\
LoRA+ epoch=1 & 42.96 \\
LoRA+ epoch=2 & 52.89 \\
LoRA+ epoch=4 & 47.65
\\\hline

\end{tabular}
\caption{Finetuned Qwen1.5-4B-Chat with different methods. Reducing learning rate can only mitigate the decline in accuracy, but it does not lead to an precision improvement.}
\label{tab:llm_4B}
\end{table}

\section{Conclusion and Limits}

\begin{figure}[ht]
\centering
\includegraphics[width=0.8\textwidth]{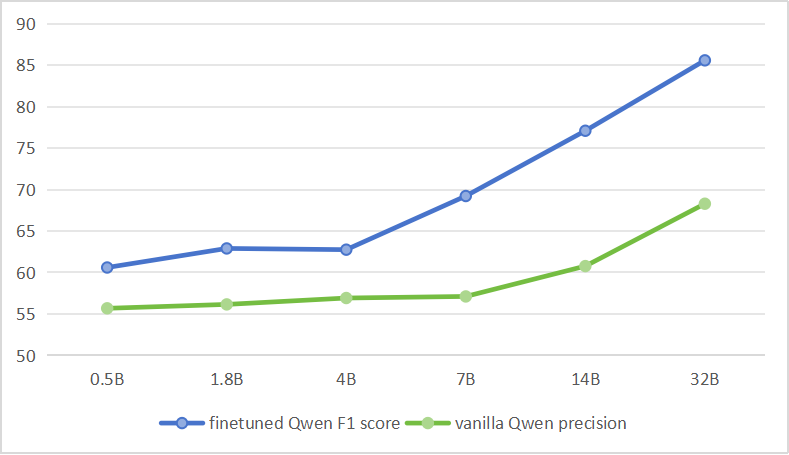}
\caption{\label{fig:overall}We compared the performance of the vanilla Qwen series with those fine-tuned models. The results showed that as the model parameters increased, the F1 scores and precision improved, which further confirmed the reliability of our dataset.}
\end{figure}

We take scaling law as guiding principle to define the problem and manually annotate the data. The model after SFT also conforms to the laws of scaling as shown in Figure \ref{fig:overall}, which proves the effectiveness of annotation.

However, Qwen1.5-4B-Chat issue remains unresolved. We need to further prove that LoRA prerequisites are met. As a full-stack open-sourced project, HuixiangDou also needs to integrate the model and verify its overall effectiveness.

\newpage
\bibliography{iclr2024_conference}
\bibliographystyle{plainnat}
\end{document}